# Efficient Euclidean Projections onto the Intersection of Norm Balls

Hao Su[*][1]  HAOSU@CS.STANFORD.EDU
Adams Wei Yu[*][2]  WYU@CS.HKU.HK
Li Fei-Fei[1]  FEIFEILI@CS.STANFORD.EDU
1 Computer Science Department, Stanford University, Stanford, CA
2 Department of Computer Science, The University of Hong Kong, Hong Kong

## Abstract

Using sparse-inducing norms to learn robust models has received increasing attention from many fields for its attractive properties. Projection-based methods have been widely applied to learning tasks constrained by such norms. As a key building block of these methods, an efficient operator for Euclidean projection onto the intersection of $\ell_1$ and $\ell_{1,q}$ norm balls ($q=2$ or $\infty$) is proposed in this paper. We prove that the projection can be reduced to finding the root of an auxiliary function which is piecewise smooth and monotonic. Hence, a bisection algorithm is sufficient to solve the problem. We show that the time complexity of our solution is $O(n + g \log g)$ for $q = 2$ and $O(n \log n)$ for $q = \infty$, where $n$ is the dimensionality of the vector to be projected and $g$ is the number of disjoint groups; we confirm this complexity by experimentation. Empirical study reveals that our method achieves significantly better performance than classical methods in terms of running time and memory usage. We further show that embedded with our efficient projection operator, projection-based algorithms can solve regression problems with composite norm constraints more efficiently than other methods and give superior accuracy.

## 1. Introduction

The sparse-inducing norms have been powerful tools for learning robust models with limited data in high-dimensional space. By imposing such norms as the constraints to the optimization task, one could bias the model towards learning sparse solutions, which in many cases has been proven to be statistically effective. Typical sparse-inducing norms include the $\ell_1$ norm and $\ell_{1,q}$ norm (defined in Sec 3) (Liu & Ye, 2010); the former encourages element-wise sparsity and the latter encourages group-wise sparsity. In a variety of contexts, the two types of sparsity pattern exist simultaneously. For example, in the multi-task learning setting, the index set for features of each task may be sparse and there might be a large overlap of features across multiple tasks (Jalali et al., 2010). One natural approach is to formalize an optimization problem constrained by $\ell_1$ norm and $\ell_{1,q}$ norm together, so that $\ell_1$ norm induces the sparsity in features of each task and $\ell_{1,q}$ norm couples the sparsity across tasks(Friedman et al., 2010).

Projection-based algorithms such as Projected Gradient (Bertsekas, 1999), Nesterov's optimal first-order method (Beck & Teboulle, 2009; Nesterov, 2007) and Projected Quasi-Newton (Schmidt et al., 2009) are major approaches to minimize a convex function with constraints. These algorithms minimize the objective function iteratively. By invoking the projection operation, the point of each iteration is guaranteed to be in the constraint set. Therefore, the projection serves as a key building block for such type of method.

In this paper, we study the problem of projecting a point in a high-dimensional space into the constraint set formed by the $\ell_1$ and $\ell_{1,q}$ norms simultaneously, in particular, $q = 2$ or $\infty$. We choose $q = 2$ or $\infty$ because these two types of norms are the most widely used group-sparsity inducing norms(Bach et al., 2011). Our Euclidean projection operator $\mathcal{P}^{\tau_1,\tau_2}_{(1,q)+1}(\mathbf{c})$ can be formulated as

$$\mathcal{P}^{\tau_1,\tau_2}_{(1,q)+1}(\mathbf{c}) = \arg\min_{\mathbf{x}}\{\|\mathbf{x} - \mathbf{c}\|_2^2 \mid \|\mathbf{x}\|_{1,q} \leq \tau_1, \|\mathbf{x}\|_1 \leq \tau_2\}. \quad (1)$$

where $\mathbf{c}$ is the point to be projected, $\|\mathbf{x}\|_{1,q}$ and $\|\mathbf{x}\|_1$

---

[*] Indicates equal contributions.

are the $\ell_{1,q}$ norm and $\ell_1$ norm of $\mathbf{x}$ (Sec 3).

We formalize the projection as a convex optimization problem and show that the solution can be parameterized by the dual variables and the parametrization has an intuitive geometrical interpretation. Since the dual problem optimizes a concave objective, intuitively we can solve the dual variables through 2-D grid search. However, this method is costly and inaccurate. Further inspection reveals that seeking the optimal dual variable can be associated with finding the unique root of a 1-D auxiliary function. As a main result of this paper, we prove that this function is piecewise smooth and strictly monotonic. Therefore, it is sufficient to adopt a bisection algorithm to handle it efficiently. We then theoretically analyze its time complexity, which are $O(n + g \log g)$ for $q = 2$ and $O(n \log n)$ for $q = \infty$.

Having obtained an efficient projection algorithm, we can embed it in the projection-based optimization methods to efficiently find the "simultaneous sparse" solution to the following problem:

$$\min_{\mathbf{w}} \; f(\mathbf{w}) \quad \text{s.t.} \quad \|\mathbf{w}\|_{1,q} \leq \tau_1, \; \|\mathbf{w}\|_1 \leq \tau_2 \quad (2)$$

where $f(\mathbf{w})$ is a convex function. We illustrate this point by experimentally solving regression problems with the above constraints.

The main contribution of this paper can be summarized as follows. Firstly, we are the first to propose a specific method that is highly efficient in time and memory usage for the composite norm projection problem. Secondly, we derive a bound to the time complexity of our algorithm and theoretically show that the algorithm enjoys fast convergence rate. This result is supported by the experiments using synthetic data and real data.

## 2. Related Work

There has been a lot of research on efficient projection operators to norm balls. Projection onto the $\ell_2$ ball is straightforward since we only need to rescale the point towards the origin. Linear time projection algorithms for $\ell_1$ norm and $\ell_{1,2}$ norm are proposed by (Duchi et al., 2008; Liu & Ye, 2009) and (Schmidt et al., 2009), respectively. For the $\ell_{1,\infty}$ norm, (Quattoni et al., 2009) proposes a method with $O(n \log n)$ complexity and (Sra, 2010) introduces a method with weak linear time complexity.

The problem of projecting a point onto the intersection of convex sets has been studied since decades ago. In particular, various alternating direction methods have been proposed(Han, 1988; Perkins, 2002; Schmidt & Murphy, 2010). For example, Dykstra's algorithm (Dykstra, 1983) and ADMM (Gabay & Mercier, 1976) are variants of the alternating projection algorithm which successively projects the point onto the convex sets until convergence. Another approach to solve the projection problem is by modeling it as a Second-Order Cone Programming (SOCP) problem and solve it by the Interior Point solver. Although these algorithms could be applied, empirical results reveal their slow convergence rate and poor scalability. Recently, (Gong et al., 2011) solves the projection onto the Intersection of Hyperplane and a Halfspace by PRF method in linear time via root finding in 1-D space. However, the problem is quite different from ours and their method could not be trivially applied. Consequently, a specific method tailored for our problem is needed.

## 3. Notations and Definitions

We start by introducing the notation and definitions that will be used throughout the paper.

Given $\mathbf{c} \in \mathbb{R}^n$, the $\ell_1, \ell_2$ and $\ell_\infty$ norms of $\mathbf{c}$ are defined as $\|\mathbf{c}\|_1 = \sum_{i=1}^n |c_i|$, $\|\mathbf{c}\|_2 = \sqrt{\sum_{i=1}^n c_i^2}$ and $\|\mathbf{c}\|_\infty = \max_{1 \leq i \leq n}\{|c_i|\}$ respectively.

In addition, the indices of $\mathbf{c}$ are divided into $g$ disjoint groups. Thus $\mathbf{c}$ can be written as $\mathbf{c} = [\widetilde{\mathbf{c}}_1; \ldots; \widetilde{\mathbf{c}}_g]$, where each $\widetilde{\mathbf{c}}_i$ is a subvector of $\mathbf{c}$. We define the $\ell_{1,q}$-norm of $\mathbf{c}$ as $\|\mathbf{c}\|_{1,q} \equiv \sum_{i=1}^g \|\widetilde{\mathbf{c}}_i\|_q$. The $(\ell_1 + \ell_{1,q})$-norm ball is defined as the intersection of $\ell_1$-norm ball and $\ell_{1,q}$-norm ball.

The Euclidean projections onto the $\ell_1, \ell_{1,q}$ and $(\ell_1 + \ell_{1,q})$ norm balls are denoted as $\mathcal{P}_1^{\tau_2}(\cdot), \mathcal{P}_{(1,q)}^{\tau_1}(\cdot)$ and $\mathcal{P}_{(1,q)+1}^{\tau_1,\tau_2}(\cdot)$, respectively.

Finally, we introduce three functions, $\text{SGN}(\mathbf{c}) = \frac{\mathbf{c}}{\|\mathbf{c}\|_2}$, $\text{MAX}(\mathbf{c}, \mathbf{d}) = [\max(c_1, d_1); \ldots; \max(c_n, d_n)]$, and $\text{MIN}(\mathbf{c}, \mathbf{d}) = [\min(c_1, d_1); \ldots; \min(c_n, d_n)]$, where $c_i, d_i$ is the $i$th element of $\mathbf{c}$ and $\mathbf{d}$.

## 4. Euclidean Projection on the $(\ell_1 + \ell_{1,q})$-norm Ball

In this section, we will introduce our approach of Euclidean projection onto the $(\ell_1 + \ell_{1,2})$-norm ball and $(\ell_1 + \ell_{1,\infty})$-norm ball. Due to space constraints, we leave most proofs in the appendix except Theorem 1.

### 4.1. Euclidean Projection on the $(\ell_1 + \ell_{1,2})$-norm Ball

In this section, we first formulate the projection on the $(\ell_1 + \ell_{1,2})$-norm ball as a convex optimization problem (Sec 4.1.1). We then parameterize the solution by dual variables and provide an intuitive geometrical interpretation (Sec 4.1.2). Finally, we determine the optimal dual variable values by finding the unique zero point of a monotonic function with a bisection algorithm (Sec 4.1.3).

#### 4.1.1. Problem Formulation

The Euclidean projection in $\mathbb{R}^n$ for the $(\ell_1+\ell_{1,2})$-norm ball can be formalized as

$$\min_{\mathbf{x}} \quad \frac{1}{2}\|\mathbf{c} - \mathbf{x}\|_2^2 \quad \text{s.t.} \quad \|\mathbf{x}\|_{1,2} \leq \tau_1, \ \|\mathbf{x}\|_1 \leq \tau_2 \quad (3)$$

where $\tau_1$ ($\tau_2$) specifies the radius of $\ell_{1,2}$ ($\ell_1$) norm ball.

Using the following proposition, we can reflect the point $c$ to the positive orthant by simply setting $c_i := |c_i|$ and later recover the original optimizer by setting $x_i^* := sign(c_i) \cdot x_i^*$. Therefore, we simply assume $c_i \geq 0, i = 1, 2, ..., n$ from now on.

**Proposition 1** Let $\mathbf{x}^*$ be the optimizer of problem (3), then $x_i^* c_i \geq 0, i = 1, 2, ..., n$.

#### 4.1.2. Parameterizing the Solution by Optimal Dual Variables

We can parameterize the solution $\mathbf{x}^*$ by the optimal dual variables $\lambda_1^*$ and $\lambda_2^*$ as shown in the following lemma, so that the KKT system (Sec A.1 in the appendix) is satisfied (Bach et al., 2011).

**Proposition 2** Suppose $\mathbf{x}^*$, $\lambda_1^*$ and $\lambda_2^*$ are the primal and dual solutions respectively, then

$$\widetilde{\mathbf{x}}_k^* = SGN(MAX(\widetilde{\mathbf{c}}_k - \lambda_2^* \widetilde{\mathbf{e}}_k, \widetilde{\mathbf{0}}_k)) \\ \cdot \max(\|MAX(\widetilde{\mathbf{c}}_k - \lambda_2^* \widetilde{\mathbf{e}}_k, \widetilde{\mathbf{0}}_k)\|_2 - \lambda_1^*, 0) \quad (4)$$

where $\widetilde{\mathbf{e}}_k$ is a vector of all $1$s which has the same dimension as $\widetilde{\mathbf{c}}_k$.

The solution has an intuitive geometrical interpretation. $\widetilde{\mathbf{x}}_k^*$ is obtained by first translating $\widetilde{c}_k$ by $MIN(\widetilde{c}_k, \lambda_2^* \widetilde{\mathbf{e}})$ units towards the origin and then shrinking by a factor of $\max(\|MAX(\widetilde{\mathbf{c}}_k - \lambda_2^* \widetilde{\mathbf{e}}_k, \widetilde{\mathbf{0}}_k)\|_2 - \lambda_1^*, 0)$. The geometrical interpretation is illustrated for the simple case where $n = 2, g = 1$ in Figure 1. According to Proposition 1, it is sufficient to consider the projection in the positive orthant. We divide the region outside the constraint set into three sets (Region I, II and III in Figure 1). The projection in Region I corresponds to the degenerated case when $\lambda_2^* = 0$ and thus $\mathbf{x}^* = \mathcal{P}_1^{\tau_2}(\mathbf{c})$ ($A_1$ is projected to $B_1$). The projection in Region II corresponds to the degenerated case when $\lambda_1^* = 0$, and thus $\mathbf{x}^* = \mathcal{P}_{1,2}^{\tau_1}(\mathbf{c})$ ($A_2$ is projected to $B_2$). The projection in Region III corresponds to the case when $\lambda_1^* > 0$ and $\lambda_2^* > 0$, where we should employ $\mathcal{P}_{(1,2),1}$ ($A_3$ is projected to $B_3$). In this simple setting with only one group, we assume $c_i - \lambda_2^* > 0, i = 1, 2$, $\|\widetilde{\mathbf{c}}_1 - \lambda_2^* \widetilde{\mathbf{e}}_1\|_2 > \lambda_1^*$, and thus (4) is reduced to $\widetilde{\mathbf{x}}_1^* = SGN(\widetilde{\mathbf{c}}_1 - \lambda_2^* \widetilde{\mathbf{e}}_1) \cdot (\|\widetilde{\mathbf{c}}_1 - \lambda_2^* \widetilde{\mathbf{e}}_1\|_2 - \lambda_1^*)$.

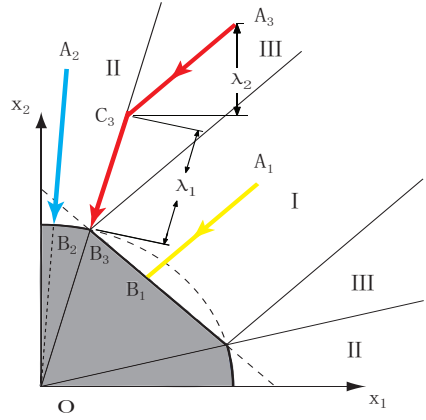

Figure 1. Geometrical interpretation for parameterizing the solution by optimal dual variables. In this simple case, we illustrate the Euclidean projection of points with only $g = 1$ group in 2-D space. Each point $A_i$ is projected to the corresponding $B_i$ through a path denoted by a bold arrow line. A quarter of the square and circle depict the $\ell_1$ and $\ell_{1,2}$ norm balls in the positive orthant respectively. The shaded area depicts the intersection of them. $\lambda_2$ stands for the vertical height of the *Translation Path* and $\lambda_1$ is the length of the *Stretch Path*.

One can find that $\widetilde{\mathbf{x}}_1^*$ ($\mathbf{OB_3}$ in Figure 1) is actually a contraction of the translated unit vector $SGN(\widetilde{\mathbf{c}}_1 - \lambda_2^* \widetilde{\mathbf{e}}_1)$ by a factor $(\|\widetilde{\mathbf{c}}_1 - \lambda_2^* \widetilde{\mathbf{e}}_1\|_2 - \lambda_1^*)$. Hence the projection path is separated into two segments. The first segment is called *Translation Path*, which is $\widetilde{\mathbf{c}}_1 - \lambda_2^* \widetilde{\mathbf{e}}_1$ ($\mathbf{A_3 C_3}$) of height $\lambda_2^*$ in Figure 1. The second segment is called *Stretch Path*, which is $\mathbf{C_3 B_3}$ of length $\lambda_1^*$ in Figure 1.

#### 4.1.3. Determining the Dual Variables

So far, we have transformed the Euclidean projection problem into determining the optimal dual variables $\lambda_1^*$ and $\lambda_2^*$. In this section, we discuss how to determine the variables case by case. We first consider the trivial cases when at least one of $\lambda_1^*$ and $\lambda_2^*$ equals to zero.

- Case 1: $\lambda_1^* = 0$ and $\lambda_2^* = 0$. This is the case when $\mathbf{c}$ is already in the $(\ell_1 + \ell_{1,2})$-norm ball (the shaded area in Figure 1) and no projection is needed. We can test this case by checking the $\ell_1$ and $\ell_{1,2}$ norms of $\mathbf{c}$.
- Case 2: $\lambda_1^* > 0$ and $\lambda_2^* = 0$. This is the case when $\mathbf{x}^* = \mathcal{P}_{1,2}^{\tau_1}(\mathbf{c})$ (Region II in Figure 1). We can test this case by checking whether $\mathcal{P}_{1,2}^{\tau_1}(\mathbf{c})$ lies in the $\ell_1$-norm ball.
- Case 3: $\lambda_1^* = 0$ and $\lambda_2^* > 0$. This is the case when $\mathbf{x}^* = \mathcal{P}_1^{\tau_2}(\mathbf{c})$ (Region I in Figure 1). We can test this case by checking whether $\mathcal{P}_1^{\tau_2}(\mathbf{c})$ lies in the $\ell_{1,2}$-norm ball.

Now we discuss the non-trivial case when $\lambda_1^* > 0$ and $\lambda_2^* > 0$ (Region III in Figure 1). According to the complementary-slackness condition of the KKT system (See (3) and (4) in the appendix), the solution satisfies

$$\sum_{i=1}^{g} \|\widetilde{\mathbf{x}}_i^*\|_2 = \tau_1, \qquad \sum_{j=1}^{n} |x_j^*| = \tau_2. \qquad (5)$$

Substitute (4) into (5) and we get the two equations of $\lambda_1$ and $\lambda_2$:

$$\tau_1 = \sum_{i \in \mathcal{S}_{\lambda_1, \lambda_2}} [\|\text{MAX}(\widetilde{\mathbf{c}}_i - \lambda_2 \widetilde{\mathbf{e}}_i, \widetilde{\mathbf{0}}_i)\|_2 - \lambda_1] \qquad (6)$$

$$\tau_2 = \sum_{i \in \mathcal{S}_{\lambda_1, \lambda_2}} (1 - \frac{\lambda_1}{\|\text{MAX}(\widetilde{\mathbf{c}}_i - \lambda_2 \widetilde{\mathbf{e}}_i, \widetilde{\mathbf{0}}_i)\|_2}) \\ \cdot \sum_{j \in \mathcal{S}_{\lambda_2}^i} (c_{i,j} - \lambda_2) \qquad (7)$$

where $\mathcal{S}_{\lambda_1,\lambda_2} = \{i \mid \|\text{MAX}(\widetilde{\mathbf{c}}_i - \lambda_2 \widetilde{\mathbf{e}}_i, \widetilde{\mathbf{0}}_i)\|_2 > \lambda_1\}$ and $\mathcal{S}_{\lambda_2}^i = \{j \mid c_{i,j} > \lambda_2\}, i = 1, 2, \ldots, g$. Now the task is to find a pair $(\lambda_1^*, \lambda_2^*)$ which satisfies (6) and (7) simultaneously.

(6) implicitly defines a function $\lambda_1(\lambda_2)$ and use this fact we obtain the following equation:

$$\lambda_1(\lambda_2) = \frac{\sum_{i \in \mathcal{S}_{\lambda_1, \lambda_2}} \|\text{MAX}(\widetilde{\mathbf{c}}_i - \lambda_2 \widetilde{\mathbf{e}}_i, \widetilde{\mathbf{0}}_i)\|_2 - \tau_1}{|\mathcal{S}_{\lambda_1, \lambda_2}|} \qquad (8)$$

Note that (8) does not define an explicit function $\lambda_1(\lambda_2)$ since $\lambda_1$ also appears on the right side of (8). For a detailed proof that $\lambda_1$ is an implicit function of $\lambda_2$, please check Lemma 2 and Lemma 3 in the appendix.

By substituting $\lambda_1(\lambda_2)$ into (7), it is easy to see that solving the equation system (6) and (7) is equivalent to finding the zero point of the following function:

$$f(\lambda_2) = \sum_{i \in \mathcal{S}_{\lambda_1(\lambda_2), \lambda_2}} (1 - \frac{\lambda_1(\lambda_2)}{\|\text{MAX}(\widetilde{\mathbf{c}}_i - \lambda_2 \widetilde{\mathbf{e}}_i, \widetilde{\mathbf{0}}_i)\|_2}) \\ \cdot \sum_{j \in \mathcal{S}_{\lambda_2}^i} (c_{i,j} - \lambda_2) - \tau_2 \qquad (9)$$

The following theorem states that $f(\lambda_2)$ is continuous, piece-wise smooth and monotone. The fact immediately leads to a bisection algorithm to efficiently find the zero point.

**Theorem 1** *1) $f$ is a continuous piece-wise smooth function in $(0, \max\{c_{i,j}\})$; 2) $f$ is monotonically decreasing and it has a unique root in $(0, \max\{c_{i,j}\})$.*

We leave the proof of the continuity and piecewise smooth property in the appendix (Lemma 5 in the appendix). Here we just prove the monotonicity of $f(\lambda_2)$.

**Proof:** Because $f(\lambda_2)$ is continuous and piecewise smooth in $(0, \max\{c_{i,j}\})$, it is sufficient to prove that $f'(\lambda_2) \leq 0$ for $\lambda_2 \in \mathbb{R}_+ \backslash \mathcal{E}$, where $\mathcal{E}$ is a set containing finite points as defined in Lemma 4 in the appendix. For such points, by Lemma 4, we can always find an interval $(a, b)$ where $\mathcal{S}_{\lambda_1, \lambda_2}$ and $\mathcal{S}_{\lambda_2}^i$ do not change, hence we can denote $\mathcal{S}_1 = \mathcal{S}_{\lambda_1, \lambda_2}$ and $\mathcal{S}_2^i = \mathcal{S}_{\lambda_2}^i$ here for simplicity.

Denote $\| \cdot \|_1^i = \sum_{j \in \mathcal{S}_2^i}(c_{i,j} - \lambda_2)$ and $\| \cdot \|_2^i = \sqrt{\sum_{j \in \mathcal{S}_2^i}(c_{i,j} - \lambda_2)^2}$. Within the interval, we assume $\| \cdot \|_2^i \geq \lambda_1(\lambda_2)$. Therefore,

$$\begin{aligned} f'(\lambda_2) &= -\sum_{i \in \mathcal{S}_1} |\mathcal{S}_2^i| + \frac{1}{|\mathcal{S}_1|} (\sum_{i \in \mathcal{S}_1} \frac{\| \cdot \|_1^i}{\| \cdot \|_2^i})^2 \\ &\quad + \lambda_1(\lambda_2) \sum_{i \in \mathcal{S}_1} \frac{|\mathcal{S}_2^i| - (\frac{\| \cdot \|_1^i}{\| \cdot \|_2^i})^2}{\| \cdot \|_2^i} \\ &\leq -\sum_{i \in \mathcal{S}_1} |\mathcal{S}_2^i| + \frac{1}{|\mathcal{S}_1|} (\sum_{i \in \mathcal{S}_1} \frac{\| \cdot \|_1^i}{\| \cdot \|_2^i})^2 \\ &\quad + \min\{\| \cdot \|_2^i\} \sum_{i \in \mathcal{S}_1} \frac{|\mathcal{S}_2^i| - (\frac{\| \cdot \|_1^i}{\| \cdot \|_2^i})^2}{\min\{\| \cdot \|_2^i\}} \\ &\leq |\mathcal{S}_1|[(\frac{1}{|\mathcal{S}_1|}\sum_{i \in \mathcal{S}_1} \frac{\| \cdot \|_1}{\| \cdot \|_2})^2 - \frac{1}{|\mathcal{S}_1|}\sum_{i \in \mathcal{S}_1}(\frac{\| \cdot \|_1}{\| \cdot \|_2})^2] \\ &\leq 0 \end{aligned}$$

Since $f(0) > 0$ and $f(\max\{c_{i,j}\}) < 0$ and $\mathcal{E}$ is finite, there exists one and only one root in $[0, \max\{c_{i,j}\}]$. $\square$

Given the theorem above, it is sufficient to apply a bisection algorithm to $f(\lambda_2)$ to find its unique root. Note that it is non-trivial to evaluate $\lambda_1(\lambda_2)$ since (8) is not a definition of the function, as we discussed before. Now we introduce an algorithm *FindLambda1* to tackle it. More specifically, we first sort all the groups $\text{MAX}(\widetilde{\mathbf{c}}_k - \lambda_2 \widetilde{\mathbf{e}}_k, \widetilde{\mathbf{0}}_k), k = 1, 2, \ldots g$ in ascending order w.r.t their $\ell_2$-norms. Then we repeatedly add the group indexes one at a time to the active group set $\mathcal{S}_{\lambda_1, \lambda_2}$, calculating the corresponding $\lambda_1$ and checking its validity. This process stops as soon as $\lambda_1$, $\lambda_2$ and $\mathcal{S}_{\lambda_1, \lambda_2}$ are all consistent.

**Complexity Analysis:** Given the tolerance $\epsilon$, the bisection projection algorithm converges after no more than $\lceil \log_2[\max_i(c_i)/\epsilon] \rceil$ outermost iterations (lines 4-21 in Algorithm 1). In each iteration, *FindLambda1* dominates the complexity. In Algorithm 2, line 4 costs $O(n)$ flops. While additional $O(n)$ flops are needed for calculating the $\ell_2$-norm of each group, the sort-

**Algorithm 1** $\ell_1 + \ell_{1,2}$ Projection

1: **Input: c**, group Index of each $c_i$, $\tau_1$, $\tau_2$, $\epsilon$.
2: **Output:** $\lambda_1$, $\lambda_2$, **x**.
3: $left = 0; right = \max\{c_{i,j}\}$;
4: **while** true **do**
5: $\quad \lambda_2 = (left + right)/2$;
6: $\quad [\lambda_1, isLambda1Found] = \text{FindLambda1}(\mathbf{c}, \lambda_2, \text{group Index of each } c_i)$;
7: $\quad$ **if** $isLambda1Found == true$ **then**
8: $\quad\quad$ Evaluate $f(\lambda_2)$ according to (9);
9: $\quad\quad$ **if** $|f(\lambda_2)| < \epsilon$ **then**
10: $\quad\quad\quad$ break;
11: $\quad\quad$ **else**
12: $\quad\quad\quad$ **if** $f(\lambda_2) < -\epsilon$ **then**
13: $\quad\quad\quad\quad$ $right = \lambda_2$;
14: $\quad\quad\quad$ **else**
15: $\quad\quad\quad\quad$ $left = \lambda_2$;
16: $\quad\quad\quad$ **end if**
17: $\quad\quad$ **end if**
18: $\quad$ **else**
19: $\quad\quad$ $right = \lambda_2$;
20: $\quad$ **end if**
21: **end while**
22: Calculate **x** according to (4);

ing in line 5 takes $O(g \log g)$ flops. Finally, the complexity of line 7 to line 14 is $O(g)$. Therefore, the overall time complexity for the projection algorithm is $\lceil \log_2[\max_i(c_i)/\epsilon] \rceil \cdot O(n + g \log g)$.

### 4.2. Euclidean Projection on the $(\ell_1 + \ell_{1,\infty})$-norm Ball

The projection onto the $(\ell_1 + \ell_{1,\infty})$-norm ball could also be addressed by a bisection algorithm. We first introduce a variable **d**, and then give an equivalent formulation of the projection problem (1) for $q = \infty$ as follows:

$$\min_{\mathbf{x},\mathbf{d}} \quad \frac{1}{2}\|\mathbf{c}-\mathbf{x}\|_2^2$$

$$\text{s.t.} \quad x_{i,j} \leq d_i (i=1,2,...g), \quad \sum_{i=1}^{g} d_i \leq \tau_1, \quad (10)$$

$$\|\mathbf{x}\|_1 \leq \tau_2, \quad x_{i,j} \geq 0, d_i \geq 0.$$

Note that the formulation above differs from (Quattoni et al., 2009) in the additional term $\|\mathbf{x}\|_1 \leq \tau_2$. Similar to Sec 4.1.2, given the KKT system (in the appendix), we can parameterize the solution $\mathbf{x}^*$ using $\mathbf{d}^*$ and optimal dual variables $\lambda_1^*$ and $\lambda_2^*$:

**Proposition 3** *Suppose $\mathbf{x}^*$, $\mathbf{d}^*$ and $\lambda_1^*$, $\lambda_2^*$ are the primal and dual solution respectively, then*

$$x_{i,j}^* = \min(\max(c_{i,j} - \lambda_2^*, 0), d_i^*),$$

$$d_i^* = \frac{\sum_{j \in \mathcal{S}_{\lambda_2^*}^{i,1}} (c_{i,j} - \lambda_2^*) - \lambda_1^*}{|\mathcal{S}_{\lambda_2^*}^{i,1}|},$$

$$\lambda_1^* = \frac{\sum_{i \in \mathcal{S}_{\lambda_2^*}} \frac{\sum_{j \in \mathcal{S}_{\lambda_2^*}^{i,1}} (c_{i,j} - \lambda_2^*) - \tau_1}{|\mathcal{S}_{\lambda_2^*}^{i,1}|}}{\sum_{i \in \mathcal{S}_{\lambda_2^*}} \frac{1}{|\mathcal{S}_{\lambda_2^*}^{i,1}|}},$$

*where* $\mathcal{S}_{\lambda_2^*}^{i} \equiv \{j | c_{i,j} - \lambda_2^* > 0\} = \mathcal{S}_{\lambda_2^*}^{i,1} \bigcup \mathcal{S}_{\lambda_2^*}^{i,2}$, $\mathcal{S}_{\lambda_2^*}^{i,1} = \{j | c_{i,j} - \lambda_2^* > d_i\}$ *and* $\mathcal{S}_{\lambda_2^*}^{i,2} = \{j | 0 < c_{i,j} - \lambda_2^* \leq d_i\}$.

The proposition above reveals that $x_{i,j}^*$, $\mathbf{d}^*$ and $\lambda_1^*$ can all be viewed as functions of $\lambda_2^*$. We can substitute the above equations into the KKT system and show that $\lambda_2^*$ is the zero point of the following function

$$h(\lambda_2) = \sum_i \sum_j \min(\max(c_{i,j} - \lambda_2, 0), d_i(\lambda_2)) - \tau_2$$

We can prove that $h(\lambda_2)$ is a strictly monotonically decreasing function:

**Theorem 2** *1) $h$ is a continuous piece-wise smooth function in $(0, \max\{c_{i,j}\})$; 2) $h$ is monotonically decreasing and it has a unique root in $(0, \max\{c_{i,j}\})$.*

**Complexity Analysis:** Based upon the above theorem, we can determine $\lambda_2^*$ using the bisection Algorithm 3. For a given $\lambda_2$, $\forall i, j, \max(c_{i,j} - \lambda_2, 0)$ is determined, (Quattoni et al., 2009) shows that $\lambda_1^*$ and $d^*$ can be solved with time complexity $O(n \log n)$. Therefore, the total time complexity of the bisection algorithm is $\lceil \log_2[\max_{i,j}(c_{i,j})/\epsilon] \rceil \cdot O(n \log n)$.

## 5. Experiments

In this section, we demonstrate the efficiency and effectiveness of the proposed projection algorithm in experiments using synthetic and real-world data. Due to the space limitation, we only show the result of $\ell_1 + \ell_{1,2}$, and the case of $\ell_1 + \ell_{1,\infty}$ is shown in the appendix.

### 5.1. Efficiency of the Proposed Projection Algorithms

We first compare our methods to Interior Point (IP) method and alternating projection methods which are also applicable in solving the Euclidean projection problem. For IP, we use a highly optimized commercial software MOSEK[1]. For alternating projection methods, as there is no algorithm specifically tailored for our problem, we compare with two widely used representative algorithms – Dykstra's algorithm and ADMM algorithm. Both algorithms generate a sequence of points whose limit is the orthogonal projection onto the intersection of convex sets [2].

---
[1] We reformulate the problem as an SOCP problem. Please refer to the appendix for the SOCP formulation.
[2] Check Sec G in Appendix for more details.

Note that all methods in the comparison apply $\mathcal{P}_1^{\tau_2}(\cdot)$ in Region I and $\mathcal{P}_{1,2}^{\tau_1}(\cdot)$ in Region II. Therefore, we first show the time cost for the shared modules $\mathcal{P}_{1,q}^{\tau_1}(\cdot)$ and $\mathcal{P}_1^{\tau_2}(\cdot)$, and then compare the different projection methods in Region III. To estimate the expected running time of each method, we estimate the volume of the three regions by Monte Carlo method with uniform sampling distribution.

We generate synthetic data with different number of groups $g$ and dimensions $n$. Specifically, each dimension of a point is sampled uniformly in $[-10^3, 10^3]$, and $\tau_1^{(2)} = 5, \tau_2^{(2)} = 6, \tau_1^{(\infty)} = 5, \tau_2^{(\infty)} = 10$. Each method is run for 10 times to calculate the average running time and standard deviation. To estimate the area of each region, we sampled 10,000 i.i.d points.

**Algorithm 2** FindLambda1

1: **Input: c**, $\lambda_2$, group Index of each $c_i$.
2: **Output:** $\lambda_1$, $isLambda1Found$.
3: $isLambda1Found = false$;
4: **for each** $i$ **do** $x_i = \max(c_i - \lambda_2, 0)$;
5: Sort the groups of $\widetilde{\mathbf{x}}$ in ascending order, w.r.t their $\ell_2$-norms;
6: $sum = 0$;
7: **for** $i = g$ **down to** 1 **do**
8: $\quad sum = sum + \|\widetilde{\mathbf{x}}_i\|_2$;
9: $\quad \lambda_1 = (sum - \tau_1)/(g - i + 1)$;
10: $\quad$ **if** ($i > 1$ **and** $\|\widetilde{\mathbf{x}}_{i-1}\|_2 < \lambda_1 \leq \|\widetilde{\mathbf{x}}_i\|_2$) **or** ($i == 1$ **and** $0 < \lambda_1 \leq \|\widetilde{\mathbf{x}}_1\|_2$) **then**
11: $\quad\quad isLambda1Found = true$;
12: $\quad\quad$ break;
13: $\quad$ **end if**
14: **end for**

**Algorithm 3** $\ell_1 + \ell_{1,\infty}$ Projection

1: **Input: c**, group Index of each $c_i$, $\tau_1$, $\tau_2$, $\epsilon$.
2: **Output:** $\lambda_2, \lambda_1, \mathbf{x}$.
3: Initialization;
4: **while** $|h(\lambda_2)| > \epsilon$ **do**
5: $\quad$ **if** $h(\lambda_2) > \epsilon$ **then**
6: $\quad\quad left = \lambda_2$;
7: $\quad$ **else**
8: $\quad\quad right = \lambda_2$;
9: $\quad$ **end if**
10: $\quad \lambda_2 = (left + right)/2$;
11: $\quad$ **for each** $i, j$ **do** $x_{i,j} = \max(c_{i,j} - \lambda_2, 0)$;
12: $\quad [\mathbf{x}, \mathbf{d}, \lambda_2, \lambda_1] = \mathcal{P}_{(1,\infty)}^{\tau_1}(\mathbf{x})$;
13: $\quad$ Evaluate $h(\lambda_2)$;
14: **end while**

In our proposed projection method, as in each step the bisection algorithm halves the range of $\lambda_2$, we stop when the range is smaller than $10^{-9}$. For Dykstra's algorithm and ADMM algorithm, we stop when the $\ell_2$-norm of two consecutive projected points falls below $10^{-9}$. For IP, we stop it if the duality gap is below $10^{-9}$. Table 1 summarizes the results for $q = 2$.

We can observe from the table that: 1) our proposed technique is significantly faster than IP and alternating projection methods (Dykstra and ADMM) in Region III; 2) our method scales well to large problems.

Compared with IP, which is the runner-up algorithm in speed, our algorithm is more memory efficient since very little extra space is needed, whereas IP introduces several groups of auxiliary variables.

Compared with Dykstra's algorithm and ADMM algorithm, our method takes much less iterations to converge. As we discussed before, the number of iterations of our algorithm is bounded by $\lceil \log_2[\max_i(c_i)/\epsilon] \rceil$. When $n = 1000$ and $g = 100$, it can be calculated that our algorithm takes no more than 30 iterations. On the other hand, empirical study shows that Dykstra's algorithm takes $692 \pm 247$ iterations to converge in Region III. A closer study also shows that Dykstra's algorithm suffers from a high variance in iterations, which may be related to the order of projections. For example, $692 \pm 247$ iterations are taken if we first project onto the $\ell_1$-norm ball, whereas $3460 \pm 565$ iterations are taken if we first project onto the $\ell_{1,2}$-norm ball.

We also analyze the volume of each region by Monte Carlo sampling method. Table 2 indicates that the probability for a point falling in Region III may be very high. Because our algorithm runs much faster than competitors in Region III, its expected running time can be much shorter in general.

### 5.2. Efficiency of the Projection-based Methods in Linear Regression with Composite Norm Constraints

In this section, we show that embedded with our proposed projection operator, various projection-based method can efficiently optimize the composite norm constrained linear regression problem. These projection-based methods significantly outperform the baseline method in speed.

We embed our projection operator into three types of projection-based optimization framework, including Projected Gradient method (PG), Nesterov's optimal first-order method (Nesterov) and Projected Quasi-Newton method (PQN).

We synthesize a small-sized data and a medium-sized data. For the small-sized data, we adopt the experiment setting in (Friedman et al., 2010). We create the coefficient vector $\mathbf{w} \in \mathbb{R}^{100}$ divided in ten blocks of ten, where the last four blocks are set to all ze-ros and the first six blocks have 10, 8, 6, 4, 2 and 1 non-zero entries respectively. All the non-zero coefficients are randomly chosen as $\pm 1$. Then we generate $N = 200$ observations $\mathbf{y} \in \mathbb{R}^{200}$ by setting $\mathbf{y} = X\mathbf{w} + \epsilon$, where $X \in \mathbb{R}^{200 \times 100}$ denotes the synthetic data points of standard Gaussian distribution with correlation 0.2 within a group and zero otherwise and $\epsilon \in \mathbb{R}^{200}$ is a Gaussian noise vector with standard deviation 4.0 in each entry. For the medium-sized data, the data generation process is similar except that the first six blocks have 100, 80, 60, 40, 20 and 10 non-zero entries respectively for 4000 observations.

| $(g, n)$ | $\ell_{1,2}$ in Region I | $\ell_1$ in Region II | $\ell_1 + \ell_{1,2}$ in Region III ||||
|---|---|---|---|---|---|---|
| | | | Dykstra | ADMM | IP | Our Algo |
| $(10, 10^2)$ | $0.158 \pm 0.039$ | $0.039 \pm 0.014$ | $397 \pm 9.00$ | $54.4 \pm 5.3$ | $9.10 \pm 0.8$ | $\mathbf{0.292 \pm 0.130}$ |
| $(10, 10^3)$ | $0.324 \pm 0.702$ | $0.067 \pm 0.024$ | $124 \pm 66.0$ | $42.2 \pm 5.4$ | $37.5 \pm 2.6$ | $\mathbf{0.451 \pm 0.065}$ |
| $(10, 10^4)$ | $1.40 \pm 0.270$ | $0.30 \pm 0.06$ | $171 \pm 108$ | $78.7 \pm 38.2$ | $383 \pm 19.8$ | $\mathbf{2.80 \pm 0.390}$ |
| $(10^2, 10^3)$ | $1.40 \pm 0.280$ | $0.067 \pm 0.024$ | $4990 \pm 3680$ | $2434 \pm 633$ | $41.1 \pm 2.20$ | $\mathbf{1.60 \pm 0.380}$ |
| $(10^2, 10^5)$ | $53.0 \pm 2.9$ | $2.57 \pm 0.35$ | $28100 \pm 6760$ | $17244 \pm 4390$ | $4460 \pm 110$ | $\mathbf{66 \pm 2}$ |
| $(10^3, 10^5)$ | $460 \pm 8$ | $2.60 \pm 0.360$ | $(1.7 \pm 0.5) \times 10^6$ | $(6.0 \pm 2.3) \times 10^5$ | $30141 \pm 1792$ | $\mathbf{492 \pm 16}$ |

Table 1. Running time (in ms) of Dykstra's algorithm, ADMM, IP, and our method. The $\ell_1$ and $\ell_{1,q}$ norm projections are shared modules for all three methods.

| $(g, n)$ | $\ell_1 + \ell_{1,2}$ | $(g, n)$ | $\ell_1 + \ell_{1,2}$ |
|---|---|---|---|
| $(10, 10^2)$ | $(0, 0.03, 0.97)$ | $(10^2, 10^3)$ | $(0, 0.01, 0.99)$ |
| $(10, 10^3)$ | $(0, 0.14, 0.86)$ | $(10^2, 10^5)$ | $(0, 0, 1)$ |
| $(10, 10^4)$ | $(0, 0.64, 0.36)$ | $(10^3, 10^5)$ | $(0, 0, 1)$ |

Table 2. Region Distribution for $\ell_1 + \ell_{1,2}$. Numbers in tuples are probabilities of uniformly generated point falling in (Region I, Region II, Region III). These probabilities are estimated by Monte Carlo simulation.

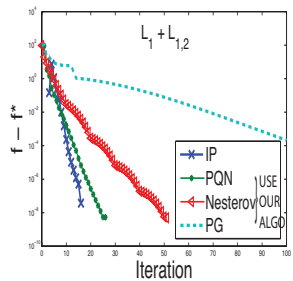

Figure 2. Number of iterations on linear regression with $(\ell_1 + \ell_{1,2})$-norm constraint for different methods. With our efficient projection operator, all three projection-based methods converge faster than IP even though they take more iterations.

Since the generated $\mathbf{w}$ exhibits sparsity in both group level and individual level, it is natural to recover $\mathbf{w}$ by solving the following constrained linear regression problem ($q = 2$ or $\infty$):

$$\min \frac{1}{2}\|\mathbf{y} - X\mathbf{w}\|_2^2 \quad \text{s.t.} \quad \|\mathbf{w}\|_{1,q} \leq \tau_1^{(q)}, \ \|\mathbf{w}\|_1 \leq \tau_2^{(q)}$$

We choose Interior Point method as the baseline to solve the above problem for its efficiency. Embedded with our efficient projection operator, all projection-based algorithms (PG, Nesterov and PQN) take much less time and memory than IP to converge to the same accuracy ($10^{-9}$) (see Table 3 for time used). We note that, projection-based methods usually take much more iterations to converge than IP (see Figure 2, and the projection operator may be invoked several times per iteration. Hence, the efficiency of the projection operator greatly impact the performance.

### 5.3. Classification Performance in Multi-task Learning

In this experiment, we show that using our efficient projection operator, with limited additional time cost, composite norm regularizer outperforms single norm regularizer in multi-task learning. In the multiple-task learning setting, there are $r > 1$ response variables (each corresponding to a task) and a common set of $p$ features. We hypothesize that, if the relevant features for each task are sparse and there is a large overlap of these relevant features across tasks, combining $\ell_{1,q}$ norm and $\ell_1$ norm will recover both the sparsity of each task and the sparsity shared across tasks.

We use handwritten digits recognition as test case. The input data are features of handwritten digits (0-9) extracted from a collection of Dutch utility maps(Asuncion & Newman, 2007). This dataset has been used by a number of papers as a reliable dataset for handwritten recognition algorithms(Jalali et al., 2010). There are $r = 10$ tasks, and each sample consists of $p = 649$ features. We use logistic regression as the classifier and constrain the classifier by $\ell_1$ norm, $\ell_{1,q}$ norm and $\ell_1 + \ell_{1,q}$ norm, respectively. PQN method is used to optimize the objective function.

We compare the running time and classification performance of each method. The classification performance is measured by the mean and standard deviation of the classification error. Results are obtained from ten ran-dom samples of training and testing data with parameters chosen via cross-validation in all methods. Using our projection operator, $(\ell_1+\ell_{1,q})$-norm yields the best classification result with similar running time to $\ell_{1,q}$-norm (Table 4). We also test replacing our projection algorithm by the runner-up algorithm in Table 1, which is IP for $q = 2$ and Dykstra for $q = \infty$. Unfortunately, using these projection operators, PQN could not converge within 30 minutes. These results show that a more structured yet complicated regularizer is more effective in a multiple-task learning problem and our efficient projection algorithms make it feasible.

| Norm Ball | $\ell_1 + \ell_{1,2}$ | | | |
|---|---|---|---|---|
| Algo<br>Size | IP | Projection-based Methods (using our projection) | | |
| | | PQN | Nesterov | PG |
| $g = 10, n = 100, \#instances = 200$ | $0.969 \pm 0.082$ | $0.121 \pm 0.007$ | $\mathbf{0.051 \pm 0.013}$ | $0.315 \pm 0.182$ |
| $g = 10, n = 2000, \#instance = 4000$ | $> 10\text{min}$ | $\mathbf{0.167 \pm 0.018}$ | $0.420 \pm 0.023$ | $0.353 \pm 0.031$ |

Table 3. Average time cost (in sec) of linear regression with $\ell_1 + \ell_{1,2}$ norm constraints. Embedded with our efficient projection operator, projected-based methods (PQN, Nesterov and PG) outperforms IP.

| | $\ell_1$ | $\ell_{1,2}$ | $\ell_{1,\infty}$ | $\ell_1 + \ell_{1,2}$ | $\ell_1 + \ell_{1,\infty}$ |
|---|---|---|---|---|---|
| Average Classification Error | 3.53% | 4.20% | 4.10% | 3.01% | **2.95%** |
| Variance of Error | 0.35% | 0.53% | 0.47% | 0.38% | 0.44% |
| Running time | $8.53s$ | $118.26s$ | $70.24s$ | $122.06s$ | $73.99s$ |

Table 4. Classification results and time cost on a Handwritten digits dataset. Using our proposed algorithm, with little additional time cost, compositing $\ell_1$ norm with $\ell_{1,q}$ norm always gives better classification performance. Our performance is on par with state-of-the-art (Jalali et al., 2010).

# References

Asuncion, Arthur and Newman, David. UCI machine learning repository, 2007.

Bach, Francis, Jenatton, Rodolphe, Mairal, Julien, and Obozinski, Guillaume. Convex optimization with sparsity-inducing norms. *Optimization for Machine Learning*, pp. 1–35, 2011.

Beck, Amir and Teboulle, Marc. Fast gradient-based algorithms for constrained total variation image denoising and deblurring problems. *IEEE Transactions on Image Processing*, 18(11):2419–2434, 2009.

Bertsekas, Dimitri P. *Nonlinear Programming*. Athena Scientific, 2nd edition, September 1999.

Duchi, John C., Shalev-Shwartz, Shai, Singer, Yoram, and Chandra, Tushar. Efficient projections onto the $l_1$-ball for learning in high dimensions. In *ICML*, pp. 272–279, 2008.

Dykstra, Richard L. An algorithm for restricted least squares regression. *JASA*, 78(384):837–842, 1983.

Friedman, Jerome, Hastie, Trevor, and Tibshirani, Robert. A note on the group lasso and a sparse group lasso. *Arxiv preprint arXiv10010736*, 2010.

Gabay, Daniel and Mercier, Bertrand. A dual algorithm for the solution of nonlinear variational problems via finite element approximation. *Computers & Mathematics with Applications*, 2(1):17–40, 1976.

Gong, Pinghua, Gai, Kun, and Zhang, Changshui. Efficient euclidean projections via piecewise root finding and its application in gradient projection. *Neurocomputing*, 74(17):2754–2766, 2011.

Han, Shih-Ping. A successive projection method. *Mathematical Programming*, 40(1):1–14, 1988.

Jalali, Ali, Ravikumar, Pradeep D., Sanghavi, Sujay, and Ruan, Chao. A dirty model for multi-task learning. In *NIPS*, pp. 964–972, 2010.

Liu, Jun and Ye, Jieping. Efficient euclidean projections in linear time. In *ICML*, pp. 83, 2009.

Liu, Jun and Ye, Jieping. Efficient $\ell_1/\ell_q$ Norm Regularization. *Arxiv preprint arXiv:1009.4766*, 2010.

Nesterov, Yu. Gradient methods for minimizing composite objective function. *ReCALL*, 76(2007076), 2007.

Perkins, Chris. A convergence analysis of dykstra's algorithm for polyhedral sets. *SIAM J. Numerical Analysis*, 40(2):792–804, 2002.

Quattoni, Ariadna, Carreras, Xavier, Collins, Michael, and Darrell, Trevor. An efficient projection for $\ell_{1,\infty}$ regularization. In *ICML*, pp. 857–864, 2009.

Schmidt, Mark W. and Murphy, Kevin P. Convex structure learning in log-linear models: Beyond pairwise potentials. In *AISTATS*, pp. 709–716, 2010.

Schmidt, Mark W., van den Berg, Ewout, Friedlander, Michael P., and Murphy, Kevin P. Optimizing costly functions with simple constraints: A limited-memory projected quasi-newton algorithm. pp. 456–463, 2009.

Sra, Suvrit. Generalized proximity and projection with norms and mixed-norms. In *Technique Report, MPI*, May 2010.